\documentclass[preprint,12pt,numbers,authoryear]{elsarticle}

\usepackage{amssymb}
\usepackage{amsmath}
\usepackage{multirow}
\usepackage{makecell} 
\usepackage{booktabs}
\usepackage{tabularx}
\usepackage{adjustbox}
\usepackage{hyperref}
\usepackage{natbib}
\usepackage{subfig}

\makeatletter
\long\def\@makecaption#1#2{%
  \vskip\abovecaptionskip\footnotesize
  \sbox\@tempboxa{#1 #2}%
  \ifdim \wd\@tempboxa >\hsize
    #1 #2\par
  \else
    \global \@minipagefalse
    \hb@xt@\hsize{\hfil\box\@tempboxa\hfil}%
  \fi
  \vskip\belowcaptionskip}
\makeatother

\makeatletter
\renewcommand{\figurename}{Fig.}
\long\def\fnum@figure{\figurename~\thefigure.}
\makeatother

\makeatletter
\long\def\fnum@table{\tablename~\thetable.}
\makeatother

\journal{Engineering Applications of Artificial Intelligence}

\begin{document}

\begin{frontmatter}



\title{Live-E2T: Real-time Threat Monitoring in Video via Deduplicated Event Reasoning and Chain-of-Thought} 

\author[bit-me]{Yuhan Wang\fnref{equ}}
\author[bit-me]{Cheng Liu\fnref{equ}}
\author[bit-me]{Zihan Zhao}
\author[bit-me]{Weichao Wu\corref{cor}}

\affiliation[bit-me]{
    organization={School of Mechatronical Engineering, Beijing Institute of Technology},
    postcode={100081},
    city={Beijing},
    country={China}}
            
\cortext[cor]{Corresponding author. E-mail: wuweichao@bit.edu.cn}
\fntext[equ]{These authors contributed equally to this work.}

\begin{abstract}
Real-time threat monitoring identifies threatening behaviors in video streams and provides reasoning and assessment of threat events through explanatory text. However, prevailing methodologies, whether based on supervised learning or generative models, struggle to concurrently satisfy the demanding requirements of real-time performance and decision explainability. To bridge this gap, we introduce Live-E2T, a novel framework that unifies these two objectives through three synergistic mechanisms. First, we deconstruct video frames into structured 'Human-Object-Interaction-Place' (HOIP) semantic tuples. This approach creates a compact, semantically focused representation, circumventing the information degradation common in conventional feature compression. Second, an efficient online event deduplication and updating mechanism is proposed to filter spatio-temporal redundancies, ensuring the system's real-time responsiveness. Finally, we fine-tune a Large Language Model (LLM) using a Chain-of-Thought (CoT) strategy, endow it with the capability for transparent and logical reasoning over event sequences to produce coherent threat assessment reports. Extensive experiments on benchmark datasets, including XD-Violence and UCF-Crime, demonstrate that Live-E2T significantly outperforms state-of-the-art methods in terms of threat detection accuracy, real-time efficiency, and the crucial dimension of explainability.
\end{abstract}



\end{frontmatter}



\section{Introduction}
\label{sec1}

Public safety monitoring systems around the world generate massive amounts of video data. Performing a timely and accurate monitoring of threatening behaviors in these data is a crucial step in maintaining social safety and preventing serious incidents \citep{Sultani_2018_CVPR}. However, existing technologies still face three core challenges when dealing with complex real-world scenarios: explainability, information fidelity, and real-time processing.

Early threat monitoring methods \citep{4407716,6531615,Hasan_2016_CVPR} relied on hand-crafted, low-level visual features, which struggled to capture high-order semantic information in complex scenes. To overcome this limitation, deep learning-based approaches \citep{Ji2022EventBasedAD,Liu2019ExploringBF,Tian2021WeaklysupervisedVA,Wu2023WeaklySA,Zhong2019GraphCL} implemented automatic learning of discriminative features through supervised end-to-end training, thus enhancing recognition accuracy. However, the highly non-linear processing of such end-to-end coupled designs renders their decision-making processes opaque, severely restricting their application in security domains that demand high levels of trust and reliability.

\begin{figure}[htpb]
    \centering
    \includegraphics[width=\textwidth]{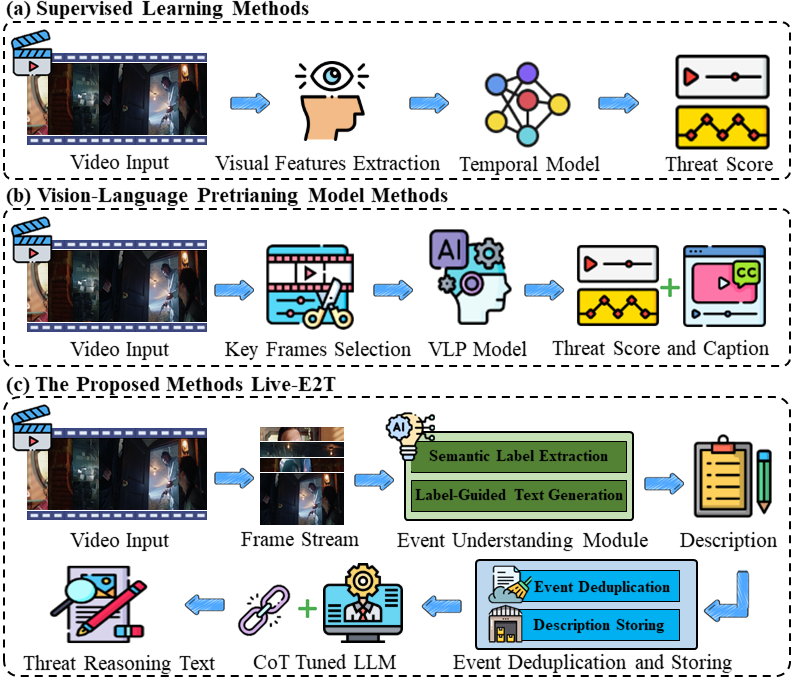}
    \caption{Research Motivation. (a) Supervised learning-based paradigm. These methods typically map video features directly to anomaly scores, but their 'black-box' decision-making process lacks explainability. (b) VLP model-based paradigm. This class of methods enhances performance by aligning multimodal semantics but often relies on offline key-frame sampling mechanisms, making it difficult to meet the demands of real-time monitoring. (c) Our proposed Live-E2T framework. Unlike existing methods, our framework constructs an end-to-end streaming pipeline. For the first time, it simultaneously achieves real-time threat monitoring and the generation of interpretable textual narratives within a unified framework, effectively overcoming the limitations of previous approaches.}
    \label{fig:1}
\end{figure}

In recent years, Vision-Language Pre-trained (VLP) models \citep{Luo2021CLIP4ClipAE,Joo2022CLIPTSACT,Zhang2024HolmesVADTU,Wu2023VadCLIPAV,Wu2024WeaklySV} have significantly enhanced the semantic understanding of video content by embedding visual information into a shared linguistic space. Some studies \citep{Zanella2024HarnessingLL,Jiang2024FromER,Tang2024HawkLT} further introduced anomaly-related instruction fine-tuning \citep{Yuan2023TowardsSV,Du2024UncoveringWW,Radford2021LearningTV} and text generation, to provide more interpretable analyses. Despite the significant progress of VLP models, their inherent information compression mechanism is prone to losing critical spatio-temporal details when processing high-dimensional video streams, leading to biased understanding or even hallucinations. More critically, the vast majority of existing state-of-the-art (SOTA) methods rely on offline processing; their high latency fails to meet the real-time response required for threat events, potentially causing missed opportunities for optimal intervention. Therefore, developing a real-time threat monitoring system that combines high efficiency, reliability, and strong explainability has become a pressing challenge in this field.

To address these challenges, this paper proposes a novel real-time interpretable threat narration framework, Live-E2T. The framework parses complex video streams into structured textual narratives in real-time through an innovative hierarchical video understanding mechanism. The core contribution of Live-E2T lies in its deconstruction of the video analysis task into two synergistic levels: precise 'frame-level' description and comprehensive 'video-level' reasoning. At the frame level, we abandon the practice of directly compressing video frames. Instead, we utilize structured semantic tags (covering human, object, action, place) to guide the model in generating precise single-frame event descriptions and innovatively use 'text-timestamp' pairs to represent the video stream, thereby effectively circumventing the information bottleneck of VLP models. 

At the video level, we design an efficient event deduplication and update mechanism. Based on the regular and repetitive nature of events in surveillance scenarios, this mechanism stores and updates descriptive text only when a new category of event appears, which significantly reduces data redundancy and guarantees the system's real-time performance. Finally, to achieve an interpretable threat assessment, Live-E2T utilizes a Large Language Model (LLM) fine-tuned with Chain-of-Thought (CoT) prompting to conduct comprehensive reasoning on the textual descriptions of historical and current events. This generates a coherent and logically clear video-level threat assessment report, providing a transparent basis for the judgment of human decision-makers. Our main contributions are as follows.

\begin{itemize}
    \item We propose Live-E2T, a novel real-time interpretable threat narration framework. Through an innovative hierarchical understanding mechanism, an event deduplication and update module, and a CoT based reasoning engine, the framework generates interpretable, real-time threat assessments from continuous video streams.
    \item We design a video representation paradigm centered on structured text. This approach transforms video into concise event descriptions, replacing traditional frame compression. Hence, it preserves critical semantics while significantly reducing data redundancy to achieve high-precision, low-latency analysis.
    \item Our comprehensive evaluations on multiple authoritative benchmark datasets validate the superior performance of Live-E2T. Experimental results demonstrate that our method significantly outperforms existing SOTA methods in terms of threat identification accuracy, multidimensional semantic understanding, and the explainability of the decision-making process.
\end{itemize}

\section{Related Works}
\label{sec2}

\subsection{Threat Monitoring based on Supervised Learning}
\label{subsec2.1}

Traditional methods for video threat monitoring relied primarily on hand-made visual features \citep{4407716,6531615,Hasan_2016_CVPR}, but these approaches exhibited limited generalizabilities in complex scenarios. With the advent of deep learning, the research focus shifted towards automatic feature extraction. Early attempts employed One-Class Support Vector Machines \citep{Ji2022EventBasedAD} to classify deep features. Zhong et al. \citep{Zhong2019GraphCL} introduced Graph Convolutional Networks to identify anomalous behaviors by constructing a spatio-temporal graph that models feature similarity and temporal consistency among video clips. To further enhance the model's capacity for capturing time-series dynamics, researchers drew upon advanced architectures from the field of Natural Language Processing. Tian et al. \citep{Tian2021WeaklysupervisedVA} proposed the Robust Temporal Feature Magnitude (RTFM) model, which utilizes a self-attention mechanism to capture long-range temporal dependencies in surveillance videos. Wu et al. \citep{Wu2023WeaklySA} developed a hierarchical learning network that integrates global and local attention modules to extract more discriminative multi-scale temporal feature embeddings.

\subsection{Threat Monitoring based on Vision-Language Models.}
\label{subsec2.2}

In recent years, VLP models, represented by CLIP \citep{Radford2021LearningTV}, have learned powerful multimodal alignment capabilities through training on large-scale image-text pairs, opening up a new paradigm for threat monitoring. CLIP4Clip \citep{Luo2021CLIP4ClipAE} pioneered the transfer of CLIP's knowledge to video-text retrieval tasks, laying the groundwork for subsequent research. Joo et al. \citep{Joo2022CLIPTSACT} introduced a temporal shift attention mechanism (CLIP-TSA) to CLIP, enhancing the model's sensitivity to temporal dependencies in videos. The VadCLIP model \citep{Wu2023VadCLIPAV}, which serves as a concise yet powerful baseline, demonstrated that image-based VLP models can be applied effectively to general-purpose threat monitoring tasks with simple adaptations. To further achieve a degree of explainability, the HolmesVAD framework designed by Zhang et al. \citep{Zhang2024HolmesVADTU} employs a lightweight temporal sampler to select key frames containing anomalous information, utilizes ImageBind \citep{Zhu2023LanguageBindEV} for multimodal encoding, and finally hands them over to a fine-tuned large multimodal model (MLLM) to generate explanatory text.

\subsection{Towards interpretable and Hierarchical Threat Monitoring.}
\label{subsec2.3}

The latest research trend has begun to focus on constructing interpretable, hierarchical video analysis frameworks. The core idea of such methods is to decompose the complex video understanding task into sub-tasks of multiple granularities, ranging from local event description to global situational reasoning. For example, LAVAD \citep{Zanella2024HarnessingLL} achieves clip-level to video-level anomaly scoring and reasoning by generating textual descriptions for video frames and designing elaborate prompts for a LLM. Zhang et al. \citep{Zhang2024HolmesVAUTL} designed an anomaly-focused temporal sampler that combines an anomaly scorer with a density-aware strategy to adaptively select video frames rich in anomaly information. This mechanism enables the subsequent MLLM to concentrate its computational resources on the most critical video regions, thus enhancing the efficiency and accuracy of the analysis. To explicitly integrate rule-based knowledge with data-driven learning within the model, Jiang et al. \citep{Jiang2024FromER} proposed RuleVM, a model employing a dual-branch architecture. Its implicit branch utilizes purely visual features for coarse-grained binary classification, while the explicit branch leverages language-image alignment techniques for more fine-grained, interpretable classification. These pioneering works collectively point to a trend: future threat monitoring systems must not only identify threats but also explain why they are threats, which is precisely the core problem our work aims to address.

\begin{figure}[htpb]
\centering
\includegraphics[width=\textwidth]{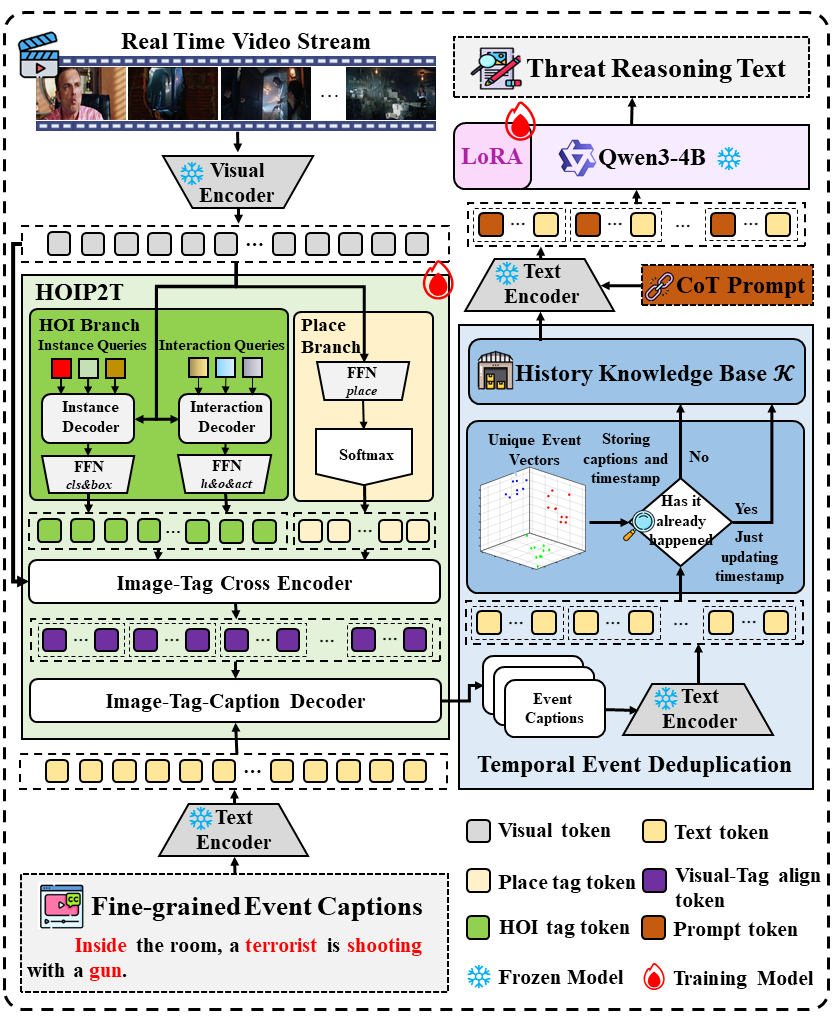}
\caption{Overview of the Live-E2T framework. The framework sequentially employs a fine-grained event understanding module (HOIP2T) to extract event semantics from the video stream, followed by a temporal event deduplication module to filter redundancies. Finally, a CoT reasoning LLM synthesizes the refined events to enable real-time and accurate video threat monitoring.}
\label{fig:2}
\end{figure}

\section{Live-E2T}
\label{sec3}
To address the core challenges faced by existing Vision-Language Models (VLMs) in video understanding tasks, such as ambiguous semantic representation, susceptibility to machine hallucination, and a lack of real-time capability, we designed and implemented Live-E2T, a hierarchical video understanding framework for real-time, interpretable threat monitoring. This framework innovatively deconstructs the complex video understanding task into three synergistic modules: fine-grained event understanding, temporal event deduplication, and interpretable threat reasoning (see Figure \ref{fig:2}).

\subsection{Model Architecture}
\label{sub3.1}

\subsubsection{Fine-grained Event Understanding.}
\label{sub3.1.1}

Research has shown that introducing a structured information pathway of "image-tag-text" can provide crucial intermediate semantic anchors for text generation, thereby effectively constraining the model's output space \citep{Huang2023Tag2TextGV}. Inspired by this, this paper proposes a label-guided multimodal generation method, HOIP2T (see the left side of Figure \ref{fig:2}). The core idea of this method is to first deconstruct a complex visual scene into a set of basic semantic tags 'Human-Object-Interaction-Place' before generating the narrative text. This set of tags collectively constitutes a structured representation of the image event, to enhance semantic alignment and relationship modeling during the generation process. HOIP2T accomplishes this through the synergy of a well-designed visual branch and a language branch.

The visual branch adopts a lightweight dual-branch parallel design, where both branches share the visual features extracted by a shared CNN backbone. The HOI branch decodes the visual features into a set of entity embeddings and action embeddings through parallel entity and action decoders, respectively. To further construct a coherent and accurate understanding of events, the model matches and associates the identified actions with their corresponding entity vectors. The action representation $z_i$ from the action decoder is fed into two Feed-Forward Networks (FFNs), $\text{FFN}_h:\mathbb{R}^d\to\mathbb{R}^d$ and $\text{FFN}_o:\mathbb{R}^d\to\mathbb{R}^d$, to point to the relevant human and object entity vectors, $v_i^h$ and $v_i^o$, where $v_i^h=\text{FFN}_h(z_i)$ and $v_i^o= \text{FFN}_o(z_i)$. For a given set of $\gamma$ action classes, two additional FFNs: ${\rm FFN}_{box}:\mathbb{R}^d\rightarrow\mathbb{R}^4$ and ${\rm FFN}_{act}:\mathbb{R}^d\rightarrow\mathbb{R}^\gamma$ are employed to re-associate the entities with actions. Concurrently, the place branch performs multi-class classification on the same visual features to rapidly identify global "place" information. Through this design, the visual branch efficiently deconstructs a complex place into a set of discrete and structured semantic elements:

\begin{equation}
    {HOIP}_i=\left\{\left[h_i,o_i,a_i, p_i\right]\right\}_{i=1}^K,
    \label{eq:1}
\end{equation}

\noindent
where $h_i$, $o_i$, $a_i$, $p_i$ are the human entities, the object entities, their behavioral mappings and the place detected by ${Image}_i$ by HOIP2T, respectively. $HOIP$ is the collection of real HOI triples. $K$ denote that HOIP2T predicts a total of $K$ four-tuples across all images.

These structured semantic labels, extracted from the visual branch, serve as semantic anchors to guide the language branch generation process. Specifically, we use a cross-attention-based encoder to semantically align and fuse the raw visual features with the semantic embeddings $HOIP$. In this mechanism, semantic embeddings act as queries to retrieve and weight the regions most relevant to the threat event from the visual features. Finally, this fused representation is fed into a decoder to generate the final explanatory text.

\noindent
\subsubsection{Temporal Event Deduplication and Storing.}
\label{sub3.1.2}

In continuous video surveillance scenarios, the raw event stream generated by the model inevitably contains a large number of semantically repetitive descriptions of normal events. This redundancy not only consumes significant storage resources but also interferes with the effective identification of key dynamic changes. To address this issue, we design an event deduplication cleaner module (see the right side of Figure \ref{fig:2}) aimed at refining the event sequence to ensure its temporal consistency and information density.

The core mechanism of this module lies in maintaining a dynamically updated event-time history knowledge base $\mathcal{K}$. For a new event $e_t$ generated in timestep $t$, its textual description $d_t$ is first mapped to a high-dimensional feature vector $v_t=\mathrm{\Phi}_{\mathrm{text}}\left(d_t\right)$ by a pre-trained text encoder $\mathrm{\Phi}_{\mathrm{text}}$. Subsequently, we calculate the cosine similarity between $v_t$ and all unique event vectors $\{{v}_{i}\ |\ e_i\ \in\mathcal{K}\}$ stored in the knowledge base $\mathcal{K}$. An event is determined to be redundant if its maximum similarity to any historical event exceeds a predefined threshold $\tau$:

\begin{equation}
    {max}_{e_i\in \mathcal{K}}{\left(\frac{v_t\cdot v_i}{\left|v_t\right|\left|v_i\right|}\right)}>\tau.
    \label{eq:2}
\end{equation}

Based on this determination, we employ a cluster-center-based representative event update strategy. Specifically, if the similarity of a new event $e_t$to one or more events in the historical knowledge base $\mathcal{K}$ exceeds the threshold $\tau$, it is assigned to a dynamic event cluster $C_k$. Instead of simply storing or discarding $e_t$ we re-evaluate the representativeness of the entire cluster. First, we compute the centroid $c_k$ of the embedding vectors of all members in the event cluster $C_k$:

\begin{equation}
    {c}_k=\frac{1}{\left|C_k\right|}\sum{v_j,\ v_j\in C_k}.
    \label{eq:3}
\end{equation}

Subsequently, within the cluster $C_k$, we identify the event ${\hat{e}}_k$ that is semantically closest to the centroid vector $c_k$ as the sole representative of the cluster. Its textual description ${\hat{d}}_k$, feature vector ${\hat{v}}_k$, and timestamp $t$are then updated in the knowledge base $\mathcal{K}$. If $e_t$ is not associated with any existing cluster, it initializes a new cluster $C_\text{new}$ and is stored in$\mathcal{K}$ as its first representative event. This approach not only significantly reduces data storage redundancy but, more importantly, yields a description composed of the most representative key events through dynamic refinement and iterative optimization. This allows for a more precise highlighting of core dynamics and anomalous behaviors in the place.

\noindent
\subsubsection{Interpretably Threat Analysis via Hierarchical Chain-of-Thought.} 
\label{sub3.1.3}

To transcend simple event detection and empower our system with robust reasoning capabilities, we address the critical challenge of inherent opacity in traditional end-to-end models. We introduce a method driven by the CoT prompting strategy, which compels the LLM to externalize its reasoning process by generating a series of intermediate logical steps before rendering a final verdict \citep{Wei2022ChainOT,Kojima2022LargeLM}. Our key innovation lies in the design of a Hierarchical CoT, a structured framework tailored for video-level threat understanding. This framework decomposes the complex reasoning task into a three-tiered cognitive hierarchy: 

\textbf{1) Relational Scene Decomposition.} This foundational layer performs basic perception and structurization. Inspired by SOTA semantic segmentation models \citep{tang2023ppmobilesegexplorefastaccurate}, it deconstructs raw visual-textual data into structured relational representations by identifying key entities (e.g., humans, objects), action associations, and place contexts. 

\textbf{2) Contextual Semantic Parsing.} Based on the structured output of the first layer, this intermediate layer performs contextualized judgment. Assesses the semantic nature of behaviors by integrating contextual cues, for example, distinguishing a brawl in a boxing ring from an assault in a park \citep{Jiang2024FromER}. This layer leverages predefined entity attributes and situational norms to infer the intent and potential threat level of observed actions. 

\textbf{3) Temporal Narrative Synthesis.} The final layer synthesizes discrete judgments from different time points into a coherent narrative \citep{ma2024eavtreventawarevideotextretrieval}. It organizes the identified events chronologically, constructing a timeline that elucidates causal links, escalations, and the overall severity of the situation.

This hierarchical decomposition strategically mimics the human cognitive process for tackling complex problems: breaking down a formidable issue into a series of manageable, interconnected sub-problems. This strategy aligns with the principle of compositionality \citep{Fodor1988ConnectionismAC} and fosters a more generalizable, human-like problem-solving paradigm \citep{Tenenbaum2018BuildingMT,Axten1973HumanPS}. By structuring the reasoning process in this manner, we not only enhance the reliability of the final output but also render the entire reasoning path transparent and auditable.

\subsection{Training of HOIP2T}
\label{sub3.2}

The training of HOIP2T is a multi-task learning process, where the total loss $\mathcal{L}_{total}$ is composed of a visual branch loss $\mathcal{L}_v$ and a language branch loss $\mathcal{L}_{LM}$. This composite objective function is designed to optimize the model's capabilities in both visual understanding and textual generation simultaneously.

\subsubsection{Visual Branch Loss}
The visual branch loss, $\mathcal{L}_v$, comprises a localization loss, an action classification loss, and a scene classification loss. Each component is chosen to address a distinct aspect of the visual recognition task.

The HOIP2T enhances the modeling capability of HOI by means of the HOI pointer mechanism. This mechanism allows for accurate correlation of behaviors with corresponding entities. The human and the object entity vectors is defined as $v^h$ and $v^o$. The human pointer and the object pointer (that is, ${c}^h$ and ${c}^o$) are obtained by Eq. (\ref{eq:4})

\begin{equation}
    \begin{aligned}
        &\hat{c}^h = \arg\max_j \text{sim}(v^h, \mu_j) \\
        &\hat{c}^o = \arg\max_j \text{sim}(v^o, \mu_j),
    \end{aligned}
    \label{eq:4}
\end{equation}

\noindent
where $\mu_j$ is the  $j^{\text{th}}$  entity representation and  $\text{sim}=u^Tv/||u||||v||$. The localization loss $\mathcal{L}_{loc}$ is defined as Eq. (\ref{eq:5}): 

\begin{equation}
    \begin{aligned}
        \mathcal{L}_{loc}(c^h,c^o,a_{\sigma})= 
        &-log\frac{\text{exp}(\text{sim}(\text{FFN}_h(a_{\sigma}),\mu_{c^h})/\tau)}{\sum_{i=1}^{K}{\text{exp}(\text{sim}(\text{FFN}_h(a_{\sigma}),\mu_k)/\tau)}} \\
        &-log\frac{\text{exp}(\text{sim}(\text{FFN}_o(a_{\sigma}),\mu_{c^0})/\tau)}{\sum_{i=1}^{K}{\text{exp}(\text{sim}(\text{FFN}_o(a_{\sigma}),\mu_k)/\tau)}},
    \end{aligned}
    \label{eq:5}
\end{equation}

\noindent
where $\sigma$ represents a matched pair of human entities and object entities, and $a_{\sigma}$ is the predicted action between them. This contrastive-style loss is crucial for the pointer mechanism, as it explicitly trains the model to generate representations for actions ($a_{\sigma}$) that are maximally similar to their corresponding ground-truth entity representations ($\mu_{c^h}$ and $\mu_{c^o}$), while being dissimilar to others. This approach is inspired by recent successes in contrastive learning for visual representation \citep{chen2020simple}, ensuring precise grounding of interactions.

The action classification loss $\mathcal{L}_{act}$ is calculated using the Binary Cross-Entropy Loss (BCELoss), as shown in Eq. (\ref{eq:6}). This choice is standard for multi-label classification tasks, as it treats each action category as an independent binary decision (present or not), which is suitable for complex scenes where multiple interactions may occur simultaneously.

\begin{equation}
    \begin{aligned}
        &\mathcal{L}_{act}(a, {\hat{a}}_{\sigma})= -\sum_{i=1}^{\gamma}[a_{i} \cdot \log({\hat{a}}_{\sigma i}) + (1 - a_{i}) \cdot \log(1 - {\hat{a}}_{\sigma i})],
    \end{aligned}
    \label{eq:6}
\end{equation}

\noindent
where $\gamma$ is the total number of interaction tag categories, and $i$ is the index for the ith category. The scene classification loss $\mathcal{L}_{place}$ is computed using the standard Cross-Entropy Loss (CELoss), as defined in Eq. (\ref{eq:7}). This is the canonical choice for multi-class, single-label classification problems like identifying a scene category, as it effectively maximizes the probability of the correct prediction \citep{jadon2020survey}.

\begin{equation}
    \begin{aligned}
        \mathcal{L}_{place}(p, {\hat{p}})= -\sum_{i=1}^{P}p_{i}\text{log}({\hat{p}}_{i}),
    \end{aligned}
    \label{eq:7}
\end{equation}

\noindent
where $P$ is the total number of place tag categories, and $j$ is the index for the jth category. 

The total loss for the visual branch, $\mathcal{L}_v$, is formulated as the sum of its constituent losses, as shown in Eq. (\ref{eq:8}).

\begin{equation}
    \begin{aligned}
        \mathcal{L}_{v}= \sum_{i=1}^{K}[\mathcal{L}_{loc}(c^h,c^o,a_{\sigma})+\mathcal{L}_{act}(a, {\hat{a}}_{\sigma})+
        \mathcal{L}_{place}(p, {\hat{p}})].
    \end{aligned}
    \label{eq:8}
\end{equation}

This summation is a common practice in multi-task learning. It allows the model to jointly optimize for three distinct but related objectives: localizing entities, classifying actions, and recognizing scenes. By combining their gradients, the model learns a shared representation that is robust and beneficial for all three sub-tasks. This approach assumes that the tasks are conditionally independent, which is a reasonable simplification that proves effective in practice. While one could introduce weighting parameters for each loss term, a direct summation (i.e., equal weighting) serves as a strong baseline and avoids extensive hyperparameter tuning \citep{kendall2018multi}..

\subsubsection{Language Branch and Total Loss}
The language branch employs an autoregressive language modeling loss, $\mathcal{L}_{LM}$, to generate descriptive text, as shown in Eq. (\ref{eq:9}). This is the standard objective for training sequence-to-sequence models, maximizing the likelihood of generating the ground-truth text token by token \citep{vaswani2017attention}.

\begin{equation}
    \begin{aligned}
        \mathcal{L}_{LM} =-E_{X\to D}[CE(x,P(x))] =-E_{X\to D}[\sum_{i=1}^Nlog(P(x_i|x_{<i}))],
    \end{aligned}
    \label{eq:9}
\end{equation}

\noindent
where $x_i$ denotes the $i^{\text{th}}$ token in the text and $N$ denotes the total number of text tokens. This mechanism ensures that the output text accurately portrays and clearly explains the threat event at the semantic level.

The final training objective of HOIP2T, $\mathcal{L}_{total}$, combines the visual and language losses to achieve robust cross-modal alignment, as defined in Eq. (\ref{eq:10}).

\begin{equation}
    \mathcal{L}_{total} = \mathcal{L}_v + \mathcal{L}_{LM}.
    \label{eq:10}
\end{equation}

This unified loss function drives the model to learn a shared embedding space where visual features are tightly aligned with their corresponding textual semantics, fulfilling the core goal of interpretable threat detection.

\subsection{Supervised Fine-tuning for Structured Reasoning}
\label{sub3.3}

To implement structured reasoning, we adopt a conventional autoregressive training paradigm, for which we constructed a bespoke instruction fine-tuning dataset. The foundation of this dataset is the annotation from the open-source, fine-grained video understanding dataset, FineVideo  \citep{Farré2024FineVideo}. This is further enhanced by a semi-automatic, human-in-the-loop Chain-of-Thought (CoT) annotation engine of our own design. The specific workflow is as follows: Initially, we utilize the GPT-4.1 model \citep{OpenAI_GPT4.1_2024} to generate preliminary CoT annotations for each sample, adhering to a pre-defined three-layer hierarchical CoT structure. Subsequently, human experts conduct a rigorous correction of these machine-generated CoTs. By referencing the ground-truth annotations from FineVideo, this step aims to eliminate any inaccuracies stemming from deficiencies in temporal modeling or from model hallucinations. Ultimately, each training instance is formulated as an input-output pair. The input consists of the historical context combined with a textual description of the current frame. The corresponding output is a complete CoT text that faithfully reproduces our three-layer reasoning framework. At the model level, we selected Qwen3-4B \citep{Yang2025Qwen3TR} as the base model. To achieve parameter-efficient fine-tuning, we employed the Low-Rank Adaptation (LoRA) technique \citep{Hu2021LoRALA}. This strategy is intended to efficiently transfer the generalized knowledge from the pre-trained model to our specific task of interpretable threat detection.

\section{Experiment}
\label{sec4}
In this section, we conduct extensive experiments to comprehensively validate and demonstrate the superior performance of Live-E2T on the task of real-time threat monitoring.

\subsection{Experiment Setup}
\label{subsec4.1}

\subsubsection{Dataset}
\label{sub4.1.1}

Our model training was conducted in two stages. First, we trained the frame-level understanding model, HOIP2T, using the TD-Hoi dataset, which we enhanced with place category annotations to support the generation of four semantic labels. Second, to guide the Large Language Model (LLM) in performing hierarchical threat reasoning, we constructed a dedicated instruction fine-tuning dataset based on the FineVideo dataset. For a fair comparison with SOTA methods, we evaluate our framework on four recognized video threat detection benchmarks: XD-Violence \citep{Wu2020NotOL}, UCF-Crime \citep{Sultani_2018_CVPR}, the Real Life Violence Situations Dataset \citep{Soliman2019ViolenceRF}, and the Large-scale Anomaly Detection dataset \citep{Wan2021AnomalyDI}. From these datasets, we curated an evaluation subset of 120 high-quality videos, removing samples from the original data that were of poor quality or lacked specific semantic content.

\subsubsection{Metrics}
\label{subsec4.1.2}
To comprehensively evaluate the Live-E2T framework, we designed metrics across two dimensions. Following relevant benchmark works \citep{Hasan_2016_CVPR,Sultani_2018_CVPR,Zanella2024HarnessingLL}, we employ the standard Area Under the Curve (AUC) and Average Precision (AP) as evaluation metrics. For the reasoning and explanation dimension, traditional text generation metrics (e.g., BLEU \citep{Papineni2002BleuAM}, CIDEr \citep{Vedantam2014CIDErCI}) cannot accurately assess semantic and logical correctness. Therefore, we adopt an evaluation paradigm based on Image-based Text Generation Performance Benchmarking. We use the Deepseek-V3-0324 model as an automated evaluator to score the explanations generated by our model on the high-quality evaluation subset across three dimensions: Correctness of Information (CoI), Behavior Mapping Accuracy (BMA), and Threat Detail Orientation (TDO).

\subsubsection{Implementation details}
\label{subsec4.1.3}
For the empirical evaluation of our proposed framework, all experiments were executed on a unified computing platform equipped with four NVIDIA A6000 GPUs. The training hyperparameters for the key modules were set as follows: As the core of fine-grained event understanding, the HOIP2T module is built upon the Hoi2Threat \citep{wang2025hoi2threatinterpretablethreatdetection} model. We trained it for 20 epochs using the AdamW optimizer with an initial learning rate of 5e-6 and a batch size of 36. For the event deduplication and cleaning task, we utilized LLaMA-Index to construct the corresponding processing module. For instruction fine-tuning, we used the Low-Rank Adaptation (LoRA) technique and trained for one epoch with the AdamW optimizer. The specific hyperparameters were configured as follows: r = 16, $\alpha$ = 32, a learning rate of 2e-5, and a batch size of 32. To test real-time monitoring performance, we simulated the use of a camera in a practical scenario by uploading test videos as a stream to a local port and monitoring that port.

\begin{table}
    \centering
    \scriptsize
    \caption{Comparison of detection performance with SOTA Threat Monitoring approaches. We include the results of interpretable and uninterpretable methods.}
    \begin{tabular}{lc|c|c|c|c}
         \hline
         \multirow{2}{*}{\textbf{Methods}} 
         & \multirow{2}{*}{\textbf{Backbone}}
         & \multirow{2}{*}{\textbf{Parameter}}
         & \multirow{2}{*}{\textbf{Real-time}} & \makecell{XD-\\Violence} & \makecell{UCF-\\Crimes} \\
         \cline{5-6}
         & & & & \textbf{AP(\%)} & \textbf{AUC(\%)} \\
         \hline
         \multicolumn{6}{c}{Uninterpretable Methods}\\
         \hline
         GODS & I3D & - & $\times$&  N/A& 70.46\\
         RTFM & I3D & - & $\times$&  77.81& 84.30\\
         HL-Net & I3D & - & $\times$&  80.00& 84.57\\
         Open VAD & ViT & - & $\times$&  66.53& 86.40\\
         CLIP-TSA & ViT & - & $\times$&  82.17& 87.58\\
         TPNG & ViT & - & $\times$&  83.68& 87.79\\
         VadCLIP & ViT & - & $\times$&  84.51& 88.02\\
         STPrompt & ViT & - & $\checkmark$& 85.22&88.08\\
         \hline
         \multicolumn{6}{c}{Interpretable Methods}\\
         \hline
         Video-LLaVA & ViT & 7B & $\times$& 50.52&55.71\\
         QwenVL-2.5 & ViT & 3B & $\times$& 62.06&75.93\\
         VTimeLLM & ViT & 7B & $\times$& 68.36&80.00\\
         LAVAD & ViT & 13B & $\times$& 62.01&80.28\\
         HolmesVAU & ViT & 7B & $\times$& 87.68&88.96\\
         HolmesVAD & ViT & 2B & $\times$& 90.67&89.51\\
         Ours (Offline) & ResNet50 & 4.5B & $\times$& \textbf{91.38}&\textbf{90.53}\\
         Ours (Realtime) & ResNet50 & 4.5B & $\checkmark$& 90.76&89.82\\
         \hline
    \end{tabular}
    \label{tab:1}
\end{table}

\begin{table*}[h]
    \centering
    \scriptsize
    \caption{Comparative analysis of threat understanding and reasoning against SOTA MLLMs and threat monitoring methods.}
    \begin{tabular}{l|ccc|ccc|ccc} 
        \hline
        \multirow{2}{*}{\textbf{Methods}} &
        \multicolumn{3}{c|}{{\makecell{Large-scale\\ Anomaly Detection}}} &
        \multicolumn{3}{c|}{{\makecell{Real Life \\ Violence Situations}}} &
        \multicolumn{3}{c}{UCF-Crimes} \\
        \cline{2-10}
        & \textbf{CoI} & \textbf{BMA} & \textbf{TDO} & \textbf{CoI} & \textbf{BMA} & \textbf{TDO} & \textbf{CoI} & \textbf{BMA} & \textbf{TDO}\\
        \hline
        Video-LLaVA & 3.60& 3.07& 2.80& 3.87& 3.27& 3.07& 3.13& 2.53& 2.33 \\
        LAVAD & 4.03& 3.40& 3.13& 4.37& 3.73& 3.50& 3.13& 2.53& 2.27\\
        QwenVL-2.5 & 4.63& 4.00& 3.73& 4.53& 3.90& 3.70& 3.83& 3.23& 2.93\\
        HolmesVAD & 4.03& 3.47& 3.23& 5.23& 4.60& 4.37& 3.67& 3.10& 2.87\\
        HolmesVAU & 4.70& 4.07& 3.80& 4.80& 4.20& 3.93& 3.63& 3.03& 2.73\\
        VTimeLLM & 5.13& 4.57& 4.13& 5.43& 4.83& 4.33& 4.63& 4.07& 3.73\\
        Ours (Offline) & \textbf{6.97}& \textbf{6.33}& \textbf{6.00} &\textbf{6.00}& \textbf{5.33}& \textbf{5.07} & \textbf{6.73}& \textbf{6.00}& \textbf{5.67}\\
        Ours (Realtime) & 6.87& 6.23& 5.90 & 5.87& 5.20& 4.93 & 6.70& 5.97& 5.63\\
        \hline
    \end{tabular}
    \label{tab:2}
\end{table*}

\subsection{Main results}
\label{subsec4.2}
To systematically evaluate the performance of Live-E2T, we conducted a comprehensive comparison with several SOTA methods, encompassing various technical paradigms such as supervised learning, vision-language pre-training, and hierarchical visual understanding. As shown in Table \ref{tab:1}, we performed a detailed comparative analysis across multiple dimensions, including the explainability, real-time processing capability and performance on the UCF-Crime and XD-Violence datasets.

The experimental results clearly demonstrate that our method surpasses all existing SOTA models on key performance metrics. Specifically, the offline version of Live-E2T achieves an AP of 91.38\% on the XD-Violence dataset and an AUC of 90.53\% on the UCF-Crime dataset. Even in the online configuration, which demands higher real-time performance, its efficacy shows only a marginal decrease yet still outperforms all other SOTA methods. This outcome strongly highlights that Live-E2T successfully addresses the high-latency limitations inherent in most high-accuracy models.

Beyond its superior detection accuracy, the core innovation of Live-E2T lies in providing a key feature that existing methods struggle to balance: explainability . To quantitatively evaluate this unique capability, we designed a novel evaluation framework. Specifically, we collected the response results of each method on the test set and submitted them to Deepseek-V3-0324 for assessment. This framework scores the model responses based on three key metrics: CoI, BMA, and TDO (on a scale of 1-10, where higher scores indicate better performance). Detailed evaluation results for all models are presented in Table \ref{tab:2}.

Live-E2T achieved the highest scores across all three metrics, fully validating its superior capabilities in threat reasoning and explanation. Compared to other general-purpose large multi-modal models or threat monitoring methods with explainability, the advantages of Live-E2T are specifically manifested in: \textbf{1)} It reliably captures key targets while effectively reducing errors such as misidentifying irrelevant objects as core subjects or omitting important targets, thus demonstrating higher recognition accuracy. \textbf{2)} It accurately establishes semantic correspondence between threat behaviors and related entities, significantly minimizing mapping deviations and ensuring that the model's understanding of the threat context is highly consistent with the actual visual semantics. \textbf{3)} The generated threat event descriptions are more nuanced, providing information-rich and specific expressions that avoid the generalized summaries common in other methods. This level of detail is particularly crucial when analyzing semantically complex threat scenarios. 

These experimental results further reveal that Live-E2T is not only a high-precision detector but also a profound scene interpreter. Its capabilities far exceed those of existing methods that rely on offline processing or have inherent deficiencies in domain knowledge \citep{Zanella2024HarnessingLL}, successfully filling the research gap in real-time, accurate, and interpretable video threat analysis.

\subsection{Analytical results}
\label{subsec4.3}

\subsubsection{Effectiveness of the Fine-grained Event Understanding Module.}
\label{subsec4.3.1}

To validate the critical role of the fine-grained event understanding module (HOIP2T) in parsing complex threat scenarios, we compared its performance against current mainstream large multi-modal models for image understanding on the TD-Hoi dataset. The evaluation was conducted using five key metrics: AUC, AP, CoI, BMA, and TDO, with the results presented in Table \ref{tab:3}. The analysis reveals that our proposed HOIP2T achieves optimal performance across all evaluation dimensions. This outcome strongly demonstrates that the module not only maintains exceptionally high threat detection accuracy but also possesses the core capability for precise semantic decomposition and fine-grained description of complex, concurrent events within frame-level images.

\begin{table}[h]
    \centering
    \caption{Ablation analysis of the quality of frame-level event descriptions.}
    \begin{tabular}{l|ccccc}
         \hline
         \textbf{Methods} & \textbf{AP}(\%) & \textbf{AUC}(\%) & \textbf{CoI} & \textbf{BMA} & \textbf{TDO} \\
         \hline
         BLIP-2&  50.00&  51.50&  3.32 &  3.10 &  2.91 \\
         LLama3.2-V&  66.65&  61.64&  2.67 &  2.59 &  2.52 \\
         MiniCPM-V&  67.50&  62.38&  3.42 &  3.34 &  3.24 \\
         Gemma3&  71.51&  76.98&  4.91 &  4.86 &  4.74 \\
         Hoi2Threat&  75.99&  72.35&  5.01 &  4.97 &  4.56 \\
         Ours&  \textbf{82.25}&   \textbf{80.94}&   \textbf{5.26} &   \textbf{5.19} &   \textbf{4.86} \\
         \hline
    \end{tabular}
    \label{tab:3}
\end{table}

\begin{table}[h]
    \centering
    \caption{Ablation analysis of the impact of event deduplication (Cleaner) and Chain-of-Thought Reasoning on text generation.}
    \begin{tabular}{l|ccc}
         \hline
         \textbf{Training Strategy} & \textbf{CoI} & \textbf{BMA} & \textbf{TDO} \\
         \hline
         HOIP2T+Qwen3 &  5.81 &  5.13 &  4.82 \\
         HOIP2T+Cleaner+Qwen3 &  5.57 &  4.89 &  4.58 \\
         HOIP2T+CoT+Qwen3 &  6.28 &  5.60 &  5.29 \\
         HOIP2T+Cleaner+CoT+ Qwen3 &   \textbf{6.60} &   \textbf{5.91} &   \textbf{5.62} \\
         \hline
    \end{tabular}
    \label{tab:4}
\end{table}

\subsubsection{Synergistic Effects of Event Deduplication and Chain-of-Thought Reasoning.}
\label{subsec4.3.2}

To investigate the impact of time-series event deduplication and CoT fine-tuning strategies on the model's advanced reasoning capabilities, we designed four experimental configurations for a comparative analysis: 1) a baseline model consisting of the HOIP2T visual encoder + Qwen3-4B; 2) the baseline model augmented with an event deduplication cleaner; 3) the baseline model fine-tuned with LoRA using a CoT instruction dataset; and 4) the full model incorporating both the event deduplication cleaner and CoT fine-tuning. The experimental results, presented in Table \ref{tab:4}, reveal a complex dependency between the deduplication module and the model's reasoning ability. For the baseline model without CoT fine-tuning, the deduplication module had a slight negative impact on performance, as it filtered out event descriptions that, while redundant, might contain latent contextual information. However, for the model enhanced with CoT fine-tuning and thus possessing stronger logical reasoning skills, the deduplication module significantly improved final decision accuracy by providing a more concise, high-signal-to-noise-ratio event sequence, thereby reducing the model's reasoning burden. This indicates that effective input information refinement must be matched with the model's intrinsic reasoning capabilities to maximize performance.

\begin{table}[h]
    \centering
    \caption{Ablation analysis of the effect of Frames Per Second (FPS) on text generation quality.}
    \begin{tabular}{c|c|ccc}
         \hline
         \multicolumn{2}{c|}{\textbf{FPS}} & \textbf{CoI} & \textbf{BMA} & \textbf{TDO} \\
         \hline
         \multirow{3}{*}{Offline} & 1.875 & 6.36 & 5.69 & 5.36 \\
          & 1.5 & 6.50 & 5.80 & 5.53 \\
          & 1.25 & \textbf{6.52} & \textbf{5.84} & \textbf{5.55} \\
         \hline
         Real-time & 1.318 & 6.51 & 5.83 & 5.53 \\
         \hline
    \end{tabular}
    \label{tab:5}
\end{table}

\subsubsection{Impact of Temporal Information Density.}
\label{subsec4.3.3}

To validate the effectiveness and robustness of our adopted "current-frame-only" online strategy and to investigate the influence of varying temporal information densities on model performance, we compared our online real-time model against offline models configured with different sampling rates. As presented in Table \ref{tab:5}, the results demonstrate a strong correlation between model performance and the temporal information density of the input frame sequence. A moderate sampling rate (1.25 FPS) proves effective in filtering high-frequency redundant information, enabling the model to concentrate on critical dynamic changes and thereby achieve optimal performance. Notably, our online real-time model, operating at 1.32 FPS, delivers performance metrics that are nearly on par with the optimal offline model. This finding provides compelling evidence that our proposed Live-E2T framework effectively meets stringent real-time processing demands without compromising the quality of analysis, thus successfully striking a balance between efficiency and accuracy.

\begin{figure}[htpb]
    \centering
    \includegraphics[width=\columnwidth]{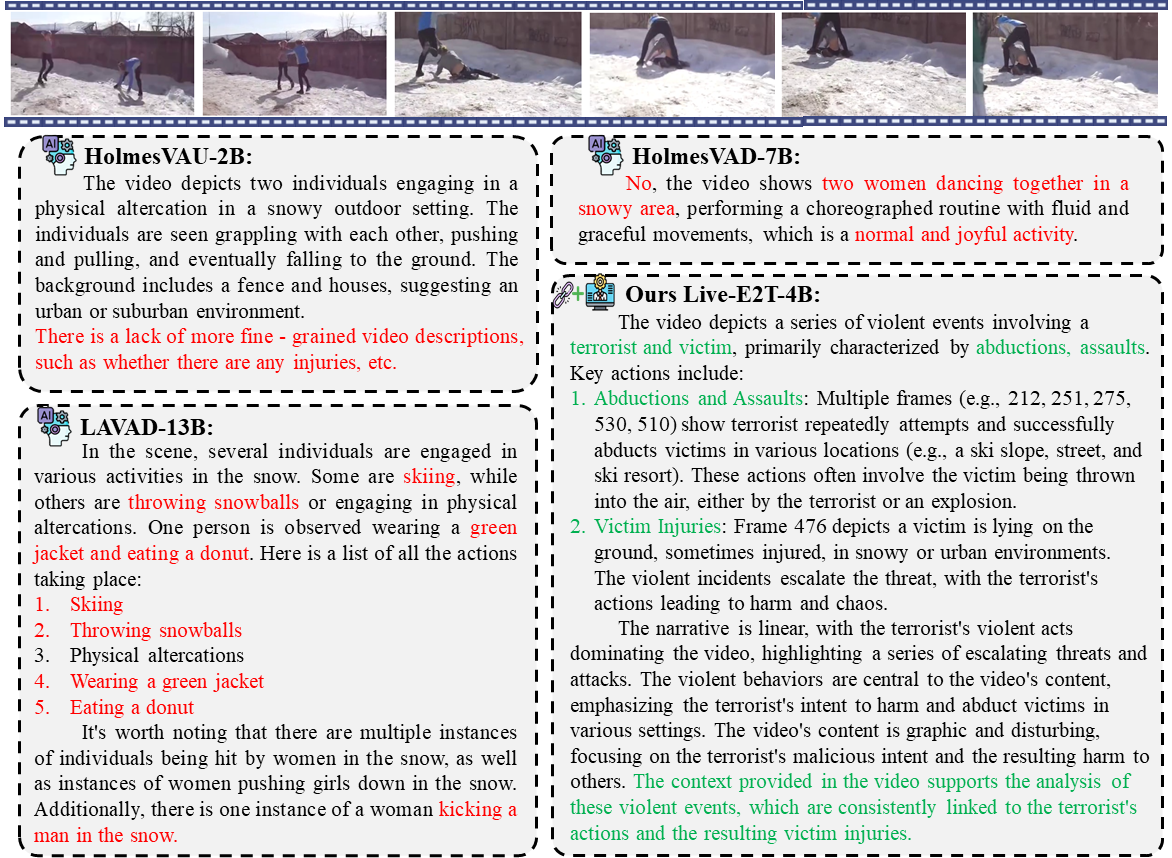}
    \caption{Qualitative comparison of anomaly understanding explanation. Compared with SOTA VAD methods(HolmesVAD, HolmesVAU, and LAVAD) in Real Life Violence Situations dataset, our proposed Live-E2T demonstrates more accurate threat judgment, along with more detailed and
 comprehensive anomaly-related descriptions and analysis. Correct and wrong explanations are highlighted in green and red, respectively.}
    \label{fig:3}
\end{figure}

\begin{figure}[htpb]
    \centering
    \includegraphics[width=\linewidth]{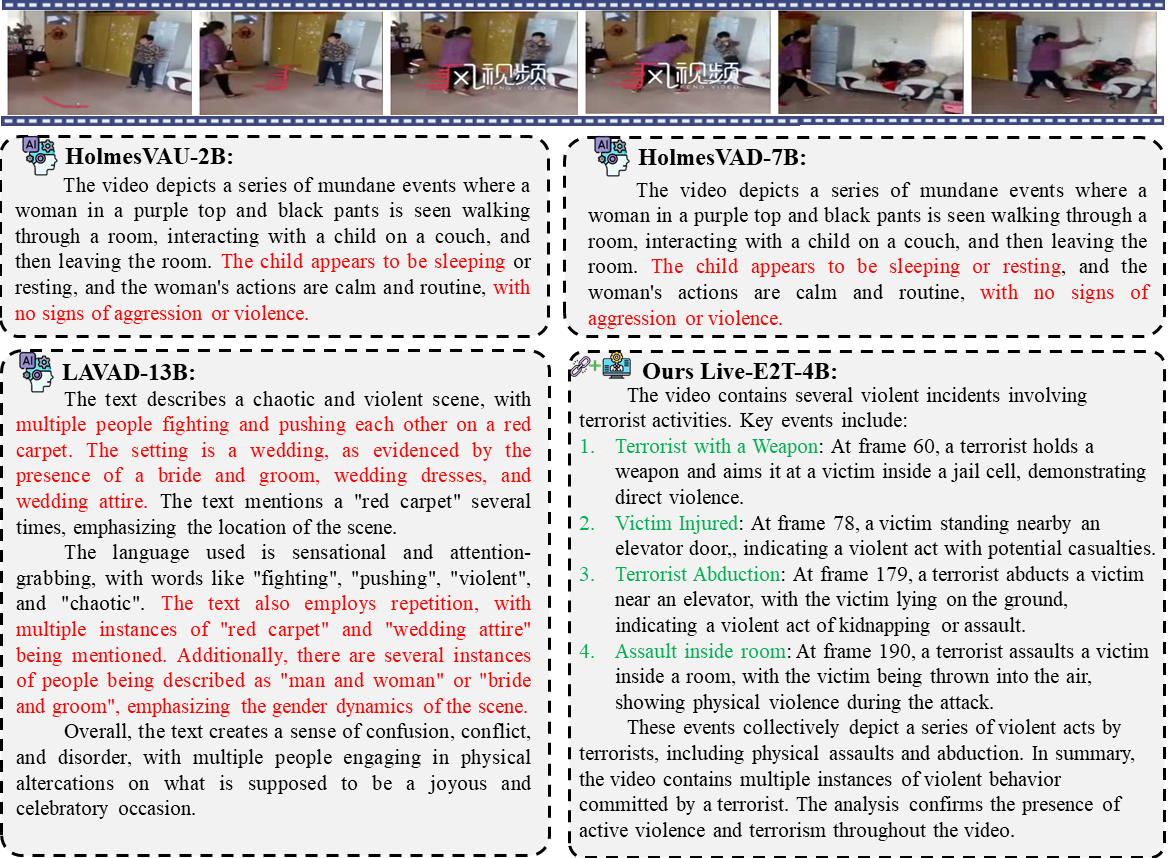}
    \caption{Qualitative comparison of threat understanding explanation with different threat monitoring methods (HolmesVAD, HolmesVAU, and LAVAD) in Large-scale Anomaly Detection dataset. Correct and wrong explanations are highlighted in green and red, respectively.}
    \label{fig:4}
\end{figure}

\begin{figure}[htpb]
    \centering
    \includegraphics[width=\linewidth]{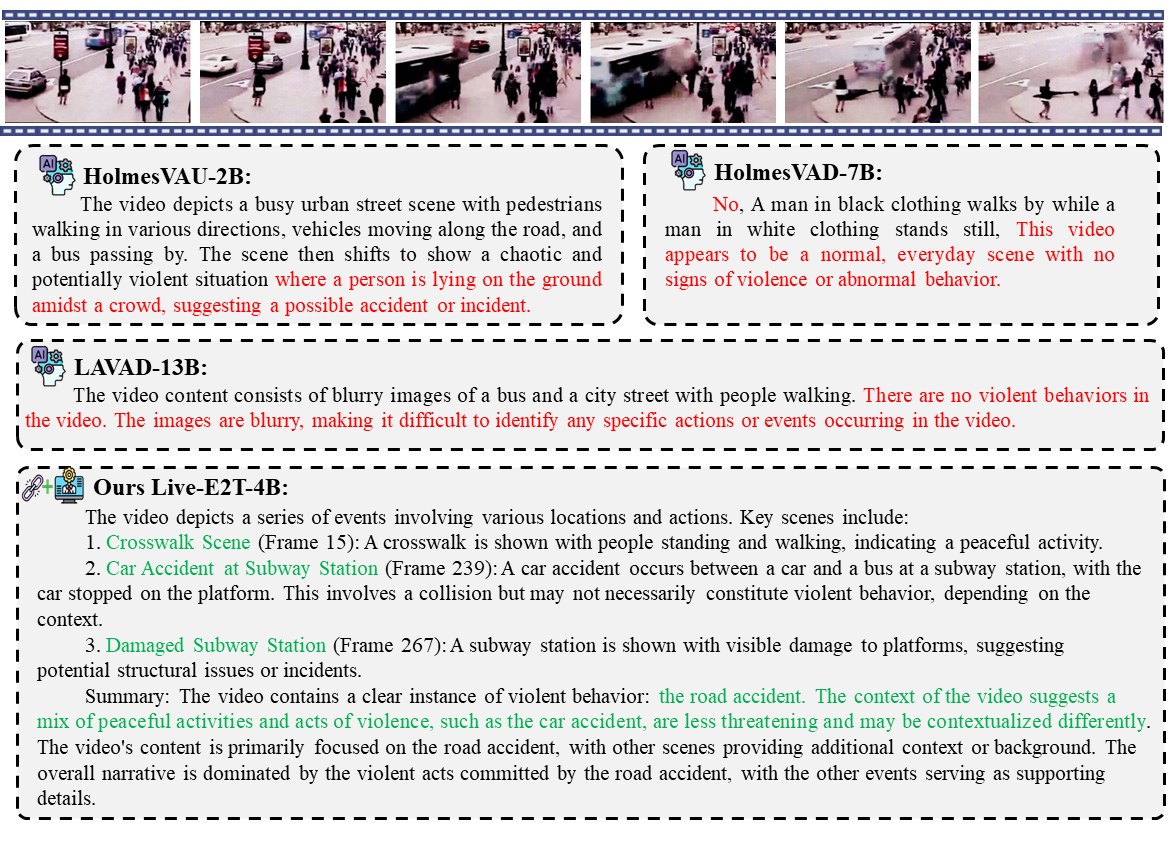}
    \caption{Qualitative comparison of threat understanding explanation with different threat monitoring methods (HolmesVAD, HolmesVAU, and LAVAD) in UCF-Crimes dataset. Correct and wrong explanations are highlighted in green and red, respectively.}
    \label{fig:5}
\end{figure}

\subsubsection{Qualitative comparison.}
\label{subsec4.3.4}

To intuitively demonstrate the superior capabilities of Live-E2T in understanding complex scenes and analyzing threats, we present a qualitative comparison with SOTA threat monitoring methods in Figure \ref{fig:3}-\ref{fig:5}. The analysis reveals that Live-E2T not only accurately identifies potential threat events within videos but also generates logically coherent and semantically consistent explanations for these threats. These results compellingly validate the effectiveness and advanced nature of Live-E2T in profoundly perceiving video dynamics and performing threat analysis.

\section{Conclusion}
\label{sec5}
In this paper, we introduce Live-E2T, a novel framework for real-time, interpretable video threat monitoring, designed to resolve the inherent trade-off between high performance and explainability in existing methods. Through an innovative architecture, Live-E2T integrates a fine-grained event understanding module (HOIP2T) for deep semantic parsing and combines it with temporal event deduplication and a CoT reasoning mechanism to achieve transparent inference from frame-level to video-level threat monitoring. Extensive experiments on benchmark datasets demonstrate that our framework significantly outperforms SOTA methods in both monitoring accuracy and its ability to generate logically coherent explanations. Our work not only establishes a new performance benchmark but also offers a viable technical pathway for developing next-generation intelligent surveillance systems that are both accurate and trustworthy.

\section*{Future work}
Although reducing video frames to textual descriptions effectively alleviates the computational pressure from real-time data streams, this process sacrifices semantic coherence over the temporal dimension. Specifically, the model faces challenges in associating semantic entities across frames, which can lead to erroneous causal inferences. Future research will be dedicated to introducing a positional encoding mechanism to enhance temporal dependency modeling capabilities. This aims to ensure the long-range consistency of historical information, thereby further improving the model's robustness and accuracy in the violence detection task.





\nocite{Wang2019GODSGO}
\nocite{Wu2023OpenVocabularyVA}
\nocite{Yang2024TextPW}
\nocite{Lin2023VideoLLaVALU}
\nocite{Huang2023VTimeLLMEL}
\nocite{Wang2024Qwen2VLEV}
\nocite{Li2023BLIP2BL}
\nocite{Yao2024MiniCPMVAG}
\nocite{Kamath2025Gemma3T}
\nocite{grattafiori2024llama3herdmodels}
\bibliography{reference} 

\end{document}